# PronouncUR: An Urdu Pronunciation Lexicon Generator

**Haris Bin Zia[1], Agha Ali Raza[1], Awais Athar[2]**

[1]Information Technology University, 6th Floor, Arfa Software Technology Park, Ferozepur Road, Lahore, Pakistan

[2]EMBL-EBI, Wellcome Genome Campus, Hinxton, Cambridgeshire, CB10 1SD, UK

{haris.zia, agha.ali.raza}@itu.edu.pk

awais@ebi.ac.uk

**Abstract**

State-of-the-art speech recognition systems rely heavily on three basic components: an acoustic model, a pronunciation lexicon and a language model. To build these components, a researcher needs linguistic as well as technical expertise, which is a barrier in low-resource domains. Techniques to construct these three components without having expert domain knowledge are in great demand. Urdu, despite having millions of speakers all over the world, is a low-resource language in terms of standard publically available linguistic resources. In this paper, we present a grapheme-to-phoneme conversion tool for Urdu that generates a pronunciation lexicon in a form suitable for use with speech recognition systems from a list of Urdu words. The tool predicts the pronunciation of words using a LSTM-based model trained on a handcrafted expert lexicon of around *39,000* words and shows an accuracy of *64%* upon internal evaluation. For external evaluation on a speech recognition task, we obtain a word error rate comparable to one achieved using a fully handcrafted expert lexicon.

## 1. Introduction

Automatic Speech Recognition (ASR) for resource scarce languages has been an active research area in the past few years (Sherwani, 2009; Qiao, 2010; Chan, 2012). Modern speech recognition systems usually require three resources: transcribed speech for acoustic modeling, a large text data for language modeling and a pronunciation lexicon that maps words to sub-word units known as phonemes. Pronunciation lexicon acts as a link connecting language model with the acoustic model.

While it is comparatively easy to gather transcribed speech waveforms and large text datasets, developing a pronunciation dictionary is quite expensive and requires tremendous amount of manual effort and linguistic expertise. Therefore, development of a pronunciation lexicon is the bottleneck when building ASR systems for low-resource languages. Techniques to reduce the need of expert knowledge in design and development of pronunciation lexicons are in great demand.

We are interested in developing a pronunciation lexicon generation tool for Urdu which is an Indo-Aryan language spoken widely with over 100 million speakers[1]. Urdu is official language of Pakistan. Its writing system is *Segmental* and more specifically *Abjad* i.e. only consonants are marked while vowels (diacritics) are optional. Urdu follows Arabic script written from right to left. A sentence written in Urdu along with its English translation is given below:

اردو پاکستان کی قومی زبان ہے ۔

Urdu is the national language of Pakistan.

Automatic Speech Recognition (ASR) research for Urdu exhibits number of challenges which are discussed in detail in subsequent sections. Despite being spoken by millions of speakers all over the world, Urdu is low-resource in terms of standard publically available linguistic resources.

To our best knowledge, our Urdu pronunciation lexicon generation tool is the first tool of its kind that makes it easier for researchers to work on Urdu speech recognition systems without prior linguistic knowledge.

The remainder of the paper is structured as follows. Section 2 reviews similar kind of work for different world languages. We then present Urdu orthography and Urdu phonetic inventory in Section 3. Section 4 briefly discusses challenges in Urdu pronunciation modeling. We present our tool in Section 5 and conclude in Section 6.

## 2. Literature Review

There exists a range of research focusing on lexical resources or tools available for different world languages for pronunciation modeling in speech recognition tasks.

- CMUdict[2] (Carnegie Mellon pronunciation dictionary) is an open-source pronunciation dictionary for North American English that contains over 134,000 words and their pronunciations (Weide, 1998). There is also a lexicon generation tool[3] available that uses CMUdict.

- Tan et al. (2009) proposed a rule based grapheme-to-phoneme tool generating a pronunciation dictionary for Malay language. Their trained ASR on read speech corpus, using tool generated pronunciation dictionary achieved a word error rate (WER) of 16.5%.

- A Bengali pronunciation dictionary[4] was developed under Google Internationalization Project[5] (Gutkin et al., 2016). The dictionary contains around 65,000 words that were manually transcribed into their phonemic representation by a team of five linguists.

[1] https://www.ethnologue.com/language/urd

[2] https://github.com/cmusphinx/cmudict

[3] http://www.speech.cs.cmu.edu/tools/lextool.html

[4] https://github.com/googlei18n/language-resources/blob/master/bn/data/lexicon.tsv

[5] https://developers.google.com/international/

- Pronunciation lexicons were developed for Amharic, Swahili and Wolof languages under LFFA Project[6] and were made available publically[7] (Gauthier et al., 2016).

- Mandarin Chinese Phonetic Segmentation and Tone is a publically[8] available corpus of 7,849 Mandarin Chinese utterances and their phonetic segmentation. The corpus can be used for pronunciation modeling of Mandarin Chinese.

- Arabic Speech Recognition Pronunciation Dictionary is a publically[9] available pronunciation dictionary for Modern Standard Arabic (MSA) that contains 526,000 words and two million pronunciations.

- Masmoudi et al. (2014) presented Tunisian Arabic Phonetic Dictionary based on a set of phonetic rules and manually tagged lexicon of exceptions (for words that do not follow phonetic rules).

- Egyptian Colloquial Arabic Lexicon is a publically[10] available pronunciation dictionary of Egyptian Colloquial Arabic (ECA), it contains 51,202 words and their pronunciation.

- The Georgetown dictionary of Iraqi-Arabic is a modern, up-to-date, publically[11] available dialectal Arabic language resource that can be used for pronunciation modeling of Iraqi-Arabic. It contains 17,500 Iraqi-Arabic entries along with their IPA pronunciations.

- Bonaventura et al. (1998) presented a letter-to-phone conversion system for Spanish that can be used to supply phonetic transcriptions to a speech recognizer.

- Mendonça et al. (2014) proposed a hybrid approach based on manual transcription rules and machine learning algorithms to build a machine readable pronunciation dictionary for Brazilian Portuguese. The dictionary as well as algorithms used to build pronunciation dictionary were made publically[12] available.

Pronunciation dictionaries developed under GlobalPhone Project (Schultz, 2014) are also available for research and commercial purposes in 20 different languages - German, French, Russian, Korean, Turkish, Chinese and Thai to name a few.

# 3. Urdu Language

## 3.1 Orthography

Urdu is written in Arabic script in a cursive format (Nastaliq style) from right to left using an extended Arabic character set. The character set includes 37 basic and 4 secondary letters, 7 diacritics, punctuation marks and special symbols (Hussain & Afzal, 2001; Afzal & Hussain, 2001; Hussain, 2004) (see Appendix A).

## 3.2 Phonetics

Urdu has a very rich phonetic inventory[13], combination of Urdu letters and diacritics realizes 44 consonants (28 non-aspirated & 16 aspirated), 7 long vowels, 7 nasalized long vowels, 3 half long vowels, 3 short vowels and 3 nasalized short vowels (Saleem et al., 2002; Hussain, 2007; Hussain, 2004). Since speech recognition systems require the representation of sounds using some phonemic notation such as IPA[14] or SAMPA[15] etc., we have used CISAMPA (Case Insensitive Speech Assessment Methods Phonetic Alphabet) proposed by Raza et al. (2010) to represent Urdu phonemes (see Appendix B).

# 4. Challenges in Urdu Pronunciation Modeling

Pronunciation modeling for Urdu exhibits a number of challenges:

*Dialects:* Due to large user base and variety of speakers, there are variations in dialect leading to large variations in pronunciation and phonetics.

*Script:* In Urdu, diacritics serve to inform reader of the short vowels accompanying each written consonant, but commonly used Urdu script generally does not contain diacritics. Speakers can distinguish the words through context and experience but some constructions may still be ambiguous, for instance, the word اس can mean either 'this' (اِس) or 'that' (اُس), their respective IPA representation being /ɪs/ or /ʊs/ respectively.

*Morphology:* Urdu is a morphologically rich language, combinations of affixes and stems results into large vocabulary of words.

*Dual Behavior:* Three Urdu characters show dual behavior i.e. both consonantal and vocalic, based on their position of occurrence (Hussain, 2004).

# 5. PronouncUR

We have developed PronouncUR, an Urdu grapheme-to-phoneme tool based on a model (c.f. Section 5.2) that can generate a pronunciation lexicon in a form suitable for use with speech recognition systems from a list of Urdu words. PronouncUR is freely available online[16].

## 5.1 Lexicon

To train our model we have developed a lexicon of approximately 46K words. Lexicon has been tagged by trained transcription experts, carefully considering the letter-to-sound rules for Urdu proposed by Hussain (2004).

---

[6] http://alffa.imag.fr/
[7] https://github.com/besacier/ALFFA_PUBLIC
[8] https://catalog.ldc.upenn.edu/LDC2015S05
[9] https://catalog.ldc.upenn.edu/LDC2017L01
[10] https://catalog.ldc.upenn.edu/LDC99L22
[11] http://press.georgetown.edu/book/languages/georgetown-dictionary-iraqi-arabic
[12] https://github.com/gustavoauma/aeiouado_g2p
[13] http://www.cle.org.pk/Downloads/ling_resources/phoneticinventory/UrduPhoneticInventory.pdf
[14] https://www.internationalphoneticassociation.org/
[15] http://www.phon.ucl.ac.uk/home/sampa/
[16] http://lextool.csalt.itu.edu.pk

The format of the training lexicon is very straight forward. Each line consists of one word form and its pronunciation. Word forms and their pronunciations are separated by tab. A small portion of the training lexicon is given in Table 1.

| | |
|---|---|
| فولاد | F O L A_A D_D |
| علامات | A L A_A M A_A T_D |
| جائیداد | D_Z A_A I_I D_D A_A D_D |
| لَڑکِیوں | L A R_R K I J O_O_N |
| درویشی | D_D A R V A_Y S_H I_I |
| الجھاؤ | U L D_Z_H A_A O_O |
| رکوا | R U K V A_A |
| ایران | I_I R A_A N |
| خریدی | X A R I_I D_D I_I |
| آفات | A_A F A_A T_D |
| فریاد | F A R J A_A D_D |
| عراقی | I R A_A Q I_I |

Table 1: Training Lexicon

Out of 67 phonemes available in Urdu phonetic inventory (see Appendix B), our training lexicon currently caters for 64 phonemes, while the work is in progress to include 3 nasalized short vowels. Phonemes M_H and J_H occur very rarely in Urdu and thus have only one entry each in the training lexicon, for the rest of the phonemes the frequency of occurrence is given in Table 2.

| # | Phoneme | Frequency | # | Phoneme | Frequency |
|---|---|---|---|---|---|
| 1 | A | 30947 | 32 | Q | 2080 |
| 2 | A_A | 27170 | 33 | X | 1641 |
| 3 | R | 18386 | 34 | R_R | 1562 |
| 4 | N | 15139 | 35 | A_Y_N | 1386 |
| 5 | I_I | 13920 | 36 | N_G | 1297 |
| 6 | I | 13683 | 37 | A_A_N | 1060 |
| 7 | L | 10909 | 38 | K_H | 1035 |
| 8 | M | 10538 | 39 | O | 928 |
| 9 | S | 10522 | 40 | G_G | 800 |
| 10 | T_D | 10075 | 41 | T_S_H | 711 |
| 11 | K | 8470 | 42 | B_H | 690 |
| 12 | A_Y | 7562 | 43 | I_I_N | 660 |
| 13 | B | 7147 | 44 | D_Z_H | 571 |
| 14 | U | 6540 | 45 | D_D_H | 555 |
| 15 | T | 6024 | 46 | T_D_H | 531 |
| 16 | D_D | 5913 | 47 | T_H | 495 |
| 17 | Z | 4940 | 48 | P_H | 435 |
| 18 | H | 4771 | 49 | G_H | 424 |
| 19 | O_O | 4766 | 50 | A_E_H | 375 |
| 20 | P | 4742 | 51 | U_U_N | 332 |
| 21 | V | 4144 | 52 | R_R_H | 225 |
| 22 | O_O_N | 4128 | 53 | D_H | 194 |
| 23 | J | 3963 | 54 | O_O_H | 70 |
| 24 | U_U | 3581 | 55 | Z_Z | 52 |
| 25 | A_E | 3440 | 56 | A_E_N | 43 |
| 26 | S_H | 3423 | 57 | Y | 36 |
| 27 | D_Z | 3331 | 58 | A_Y_H | 33 |
| 28 | G | 3275 | 59 | N_H | 12 |
| 29 | F | 3233 | 60 | L_H | 8 |
| 30 | D | 2762 | 61 | R_H | 8 |
| 31 | T_S | 2491 | 62 | O_N | 4 |

Table 2: Frequency Distribution of Phonemes in Training Lexicon

## 5.2 G2P Model

The grapheme-to-phoneme (G2P) is the task of translating input sequence of graphemes (letters) to output sequence of phonemes.

| | | | |
|---|---|---|---|
| **Graphemes** | ب | َ | ن |
| **Phonemes** | B | A | N |

Table 3: An example of grapheme-to-phoneme translation

Given the success of sequence-to-sequence learning (Sutskever et al., 2014) and power of LSTM for sequence modeling (Hochreiter et al., 1997), we choose LSTM for grapheme-to-phoneme conversion as proposed by Yao et al. (2015). We used open-source G2P toolkit[17] to train our G2P model with 2 LSTM layers and 512 hidden units in each layer.

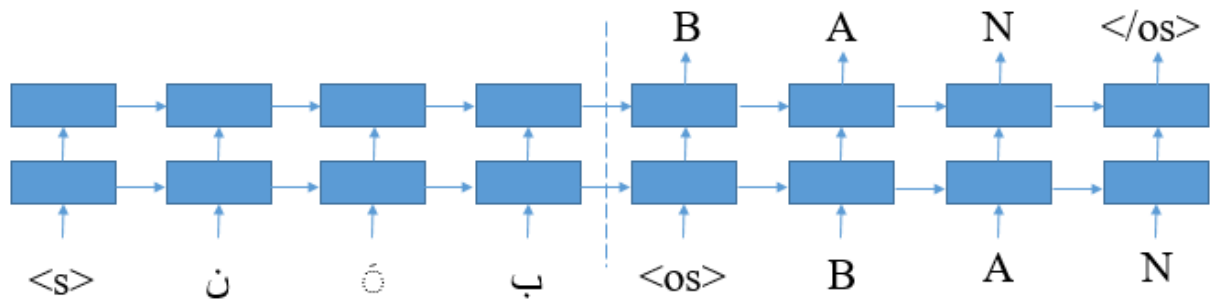

Figure 1: An encoder-decoder LSTM with two layers.

Figure 1 shows a sample of the model where the encoder LSTM is on the left of dotted line while decoder on the right. The encoder reads a time-reversed sequence “<s> ن َ ب” and produces the last hidden layer activation to initialize the decoder. The decoder reads “<os> B A N” as the past phoneme prediction sequence and uses “B A N </os>” as the output sequence to generate. <s> denotes input sequence beginning while <os> and </os> denotes output sequence beginning and ending respectively.

## 5.3 Performance Evaluation

We split our handcrafted lexicon in 85% training set, 5% validation and 10% test set. Intrinsic evaluation on unseen test set our G2P model achieved word error rate (WER) of 36%. The same G2P model trained on CMUdict has WER of 28.61% (Yao et al., 2015). The low word error rate of CMUdict can be attributed to its large size. Another reason for our comparatively higher WER may be that only about 11% of the words in our corpus have diacritics. As a result, a good performance would require overcoming the problem of automatic diacritization which gets harder while processing a list of isolated words without any context.

To perform extrinsic evaluation of the performance of lexicon tool on speech recognition task, we trained a Hidden Markov Model (HMM) based speech recognition system on phonetically rich Urdu speech corpus[18] (Raza et al., 2009) and spontaneous speech corpus (Raza et al., 2010) using CMUSphinx[19] speech recognition toolkit. The combined data from both corpora contains 3,974 utterances spanning over 179 minutes of speech, out of which 157 minutes (3,174 utterances) were used for training and 22 minutes (800 utterances) for testing. A tri-gram language model using the training data transcripts was applied during decoding. By using lexicon generated through lexicon tool, we obtained a word error rate (~19%) that approaches the rate achieved using a fully handcrafted expert lexicon. We used the same train/test split as used by Raza et al. (2010) and thus results are directly comparable.

[17] https://github.com/cmusphinx/g2p-seq2seq
[18] http://csalt.itu.edu.pk/PRUSCorpus/index.html
[19] https://cmusphinx.github.io/

## 6. Conclusion and Future Work

We presented an online pronunciation lexicon generation tool for Urdu that can be used to generate pronunciation lexicon to be used with speech recognition systems. Experimental results showed that pronunciation lexicon generated through lexicon tool behaves as good as handcrafted expert lexicon in speech recognition tasks.

As a future direction, we will look into the ways to decrease the WER of lexicon tool e.g. increase diacritic coverage in training lexicon, increase size of training lexicon, add support for nasalized short vowels and increase the coverage of rarely occurring phonemes.

## 7. Acknowledgements

We would like to thank Atique-ur-Rehman for providing us with cloud hosting and Murtaza Azam Khan for his help with frontend.

## 8. Bibliographical References

## 9. Language Resource References

## Appendix A

| | | | | | | | |
|---|---|---|---|---|---|---|---|
| چ | ج | ث | ٹ | ت | پ | ب | ا |
| ز | ڑ | ر | ذ | ڈ | د | خ | ح |
| ع | ظ | ط | ض | ص | ش | س | ژ |
| ن | م | ل | گ | ک | ق | ف | غ |
| | | | ے | ی | ء | ہ | و |

Table A1: Basic Urdu Letters

| | | | |
|---|---|---|---|
| ھ | ۃ | ں | آ |

Table A2: Secondary Urdu Letters

| | | | | | | |
|---|---|---|---|---|---|---|
| ◌ً | ◌ْ | ◌ٰ | ◌ّ | ◌ُ | ◌ِ | ◌َ |

Table A3: Urdu Diacritics

## Appendix B

| **Sr. No.** | **Urdu Letter** | **IPA** | **CISAMPA** |
|---|---|---|---|
| | **Consonants** | | |
| 1 | پ | p | P |
| 2 | پھ | pʰ | P_H |
| 3 | ب | b | B |
| 4 | بھ | bʰ | B_H |
| 5 | م | m | M |
| 6 | مھ | mʰ | M_H |
| 7 | ت،ط | t̪ | T_D |
| 8 | تھ | t̪ʰ | T_D_H |
| 9 | د | d̪ | D_D |
| 10 | دھ | d̪ʰ | D_D_H |
| 11 | ٹ | t | T |
| 12 | ٹھ | tʰ | T_H |
| 13 | ڈ | d | D |
| 14 | ڈھ | dʰ | D_H |
| 15 | ن | n | N |
| 16 | نھ | nʰ | N_H |
| 17 | ک | k | K |
| 18 | کھ | kʰ | K_H |
| 19 | گ | g | G |
| 20 | گھ | gʰ | G_H |
| 21 | ن in نک،نکھ،نگ،نگھ | ŋ | N_G |
| 22 | ق | q | Q |
| 23 | ع | ʔ | Y |
| 24 | ف | f | F |
| 25 | و | v | V |
| 26 | س | s | S |
| 27 | ذ،ز،ض،ظ | z | Z |
| 28 | ش | ʃ | S_H |
| 29 | ژ | ʒ | Z_Z |
| 30 | خ | x | X |
| 31 | غ | ɣ | G_G |
| 32 | ح،ہ | h | H |
| 33 | ل | l | L |
| 34 | لھ | lʰ | L_H |
| 35 | ر | r | R |
| 36 | رھ | rʰ | R_H |
| 37 | ڑ | ɽ | R_R |
| 38 | ڑھ | ɽʰ | R_R_H |
| 39 | ی | j | J |
| 40 | یھ | jʰ | J_H |
| 41 | چ | tʃ | T_S |
| 42 | چھ | tʃʰ | T_S_H |
| 43 | ج | dʒ | D_Z |
| 44 | جھ | dʒʰ | D_Z_H |
| | **Vowels** | | |
| 45 | ◌ُو | uː | U_U |
| 46 | و | oː | O_O |
| 47 | ◌َو | ɔː | O |
| 48 | آ،ا | aː | A_A |
| 49 | ی | iː | I_I |
| 50 | ے | eː | A_Y |
| 51 | ◌َے | æː | A_E |
| 52 | ◌ُوں | ũː | U_U_N |
| 53 | وں | õː | O_O_N |
| 54 | ◌َوں | ɔ̃ː | O_N |
| 55 | آں،اں | ãː | A_A_N |
| 56 | ◌ِیں | ĩː | I_I_N |
| 57 | یں | ẽː | A_Y_N |
| 58 | ◌َیں | æ̃ː | A_E_N |
| 59 | ◌ِہ | eˑ | A_Y_H |
| 60 | ◌َہ | æˑ | A_E_H |
| 61 | ◌ُہ | oˑ | O_O_H |
| 62 | ◌ِ | ɪ | I |
| 63 | ◌ُ | ʊ | U |
| 64 | ◌َ،ء | ə | A |
| 65 | ◌ِں | ɪ̃ | I_N |
| 66 | ◌ُں | ʊ̃ | U_N |
| 67 | ◌َں | ə̃ | A_N |

Table B1: Urdu Letters with IPA and CISAMPA